\documentclass[letterpaper]{article} 
\usepackage{aaai24}  
\usepackage{times}  
\usepackage{helvet}  
\usepackage{courier}  
\usepackage[hyphens]{url}  
\usepackage{graphicx} 
\usepackage{natbib}  
\usepackage{caption} 
\usepackage{algorithm}

\usepackage{newfloat}
\usepackage{listings}
\usepackage{xspace} 
\usepackage{algorithmicx}

\usepackage{algpseudocode} 
\usepackage{xcolor}
\newcommand{\ie}{i.e.\xspace} 
\usepackage{times}
\usepackage{epsfig}
\usepackage{amsmath}
\usepackage{amssymb}
\usepackage{multirow}
\usepackage{amsmath} 
\newcommand{\eg}[1]{\textit{e.g.} #1}

\DeclareCaptionStyle{ruled}
{labelfont=normalfont,labelsep=colon,strut=off} 
\floatstyle{ruled}
\newfloat{listing}{tb}{lst}{}
\floatname{listing}{Listing}
\title{Fine-Grained Knowledge Selection and Restoration \\for Non-Exemplar Class Incremental Learning}
\author{
    Jiang-Tian Zhai $^{1}$  \quad
    Xialei Liu $^{1, }$\thanks{Corresponding author.} \quad
    Lu Yu $^2$ \quad
    Ming-Ming Cheng $^1$ \\
    $^1$ VCIP, CS, Nankai University \quad
    $^2$  Tianjin University of Technology \quad \\
    {\tt\small jtzhai30@gmail.com, xialei@nankai.edu.cn, luyu@email.tjut.edu.cn, cmm@nankai.edu.cn}
}

\usepackage{bibentry}
\newcommand{\minisection}[1]{\vspace{0.0in} \noindent {\bf #1}}
\begin{document}

\maketitle

\begin{abstract}
Non-exemplar class incremental learning aims to learn both the new and old tasks without accessing any training data from the past. This strict restriction enlarges the difficulty of alleviating catastrophic forgetting since all techniques can only be applied to current task data. Considering this challenge, we propose a novel framework of fine-grained knowledge selection and restoration. The conventional knowledge distillation-based methods place too strict constraints on the network parameters and features to prevent forgetting, which limits the training of new tasks. To loose this constraint, we proposed a novel fine-grained selective patch-level distillation to adaptively balance plasticity and stability. Some task-agnostic patches can be used to preserve the decision boundary of the old task. While some patches containing the important foreground are favorable for learning the new task.
   Moreover, we employ a task-agnostic mechanism to generate more realistic prototypes of old tasks with the current task sample for reducing classifier bias for fine-grained knowledge restoration.  Extensive experiments on CIFAR100, TinyImageNet and ImageNet-Subset demonstrate the effectiveness of our method. Code is available at \url{https://github.com/scok30/vit-cil}. 
\end{abstract}

\section{Introduction}

Deep neural networks have achieved superior performance in many computer vision tasks. Since the real world is open and dynamic, it is important to be capable of learning new knowledge during the application of these networks. 
Incremental learning aims at learning new tasks without forgetting previous ones, and it is also a crucial characteristic of deep neural networks applied to real-world scenarios. For example, the system of face recognition may meet some faces with masks in recent years, and it is essential to adapt to these new circumstances as well. However, simply fine-tuning deep neural networks on the new task will cause severe catastrophic forgetting~\cite{robins1995catastrophic} since the network almost completely adjusts its parameters to the new task~\cite{goodfellow2013empirical,mccloskey1989catastrophic}. To address this problem, many recent works~\cite{castro2018end,douillard2020podnet,hou2019learning,liu2021adaptive,rebuffi2017icarl,yan2021dynamically} are proposed to alleviate the catastrophic forgetting problem. 

In this paper, we consider a challenging task setting of class incremental learning termed non-exemplar class incremental learning (NECIL)\cite{gao2022r,zhu2021class,zhu2022self}, that forbids the model to preserve any old task sample while learning sequential tasks with samples from disjoint classes. Compared with normal settings, non-exemplar class incremental learning considers the data privacy issue and storage burden. 
To address the issue of catastrophic forgetting in this task, researchers have introduced various approaches aimed at preserving acquired knowledge without the need of past data.

\begin{figure}[!t]
    \centering
    \includegraphics[scale=0.23]{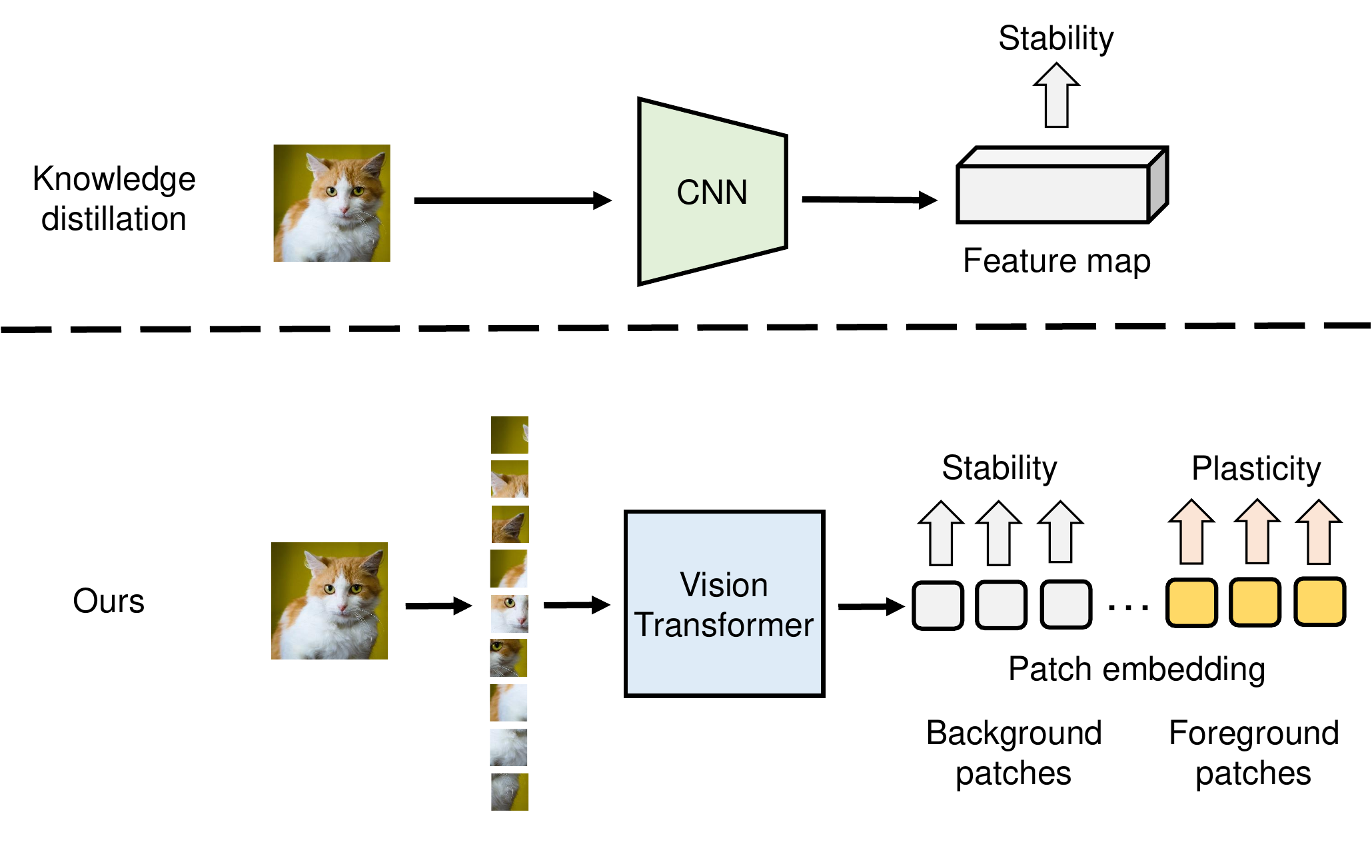}
	\caption{Comparison between conventional knowledge distillation (KD) and our patch-level fine-grained knowledge selection method based on vision Transformer architecture. Conventional KD treats the image as a whole, while patch embeddings enable us to strike a better trade-off between stability and plasticity for different local regions. Moreover, a task-agnostic fine-grained prototype restoration is proposed to better replay the old knowledge.
	}\label{fig:fig1}
 \vspace{-5mm}
\end{figure}

 Early work like LwF~\cite{li2017learning} introduces vanilla knowledge distillation between the old and current model and does not perform well under this challenging setting. Some replay-based methods usually narrow the gap by generating synthetic images. While the effectiveness of incremental learning can be influenced by the quality of the generated images. Recently, DeepInversion~\cite{yin2020dreaming} proposes to invert the trained networks from random noise to generate images as pseudo exemplars for training together with current data. R-DFCIL~\cite{gao2022r} introduces the relation-guided supervision to overcome the severe domain gap between the synthetic
and real data to generate more realistic old samples for replay. In comparison, PASS~\cite{zhu2021prototype} considers the problem of classifier bias and proposes to maintain the decision boundary of old tasks by applying augmented prototype replay. SSRE~\cite{zhu2022self} follows its design and introduces a dynamically expanded model structure for adaptation to new task knowledge while reducing the negative effect of forgetting old ones. However, most of these methods depend on a common and basic module for alleviating forgetting, \ie{ knowledge distillation~\cite{hinton2015distilling}}. 

When knowledge distillation is applied to  NECIL, it adopts the current task sample and reduces the representation distance from the old and new models. It aims to enhance the model stability when learning new tasks to reduce forgetting of old knowledge. However, this operation partially contradicts with learning the current task, especially on NECIL settings, since we expect the model to achieve plasticity on new data. The intrinsic reason 
is that the previous methods treat this distillation process from the sample level, which is less informative since only part of the region, \ie{ the foreground} in the image is highly related to the current classification task. If we can separately apply different strategies on task-relevant and task-agnostic regions, we can achieve a much more fine-grained learning process for plasticity and stability trade-offs. In other words, the model is taught to remember representations of the common background regions and learn discriminative and task-relevant knowledge of the foreground. A concept illustration can be seen in Figure~\ref{fig:fig1}. 

Another challenge of NECIL is the classifier bias across tasks described in PASS~\cite{zhu2021prototype}. The prototype is a one-dimension embedding of the input image for classification computed with the network (\eg{the result of average pooling on the feature map from CNN, or the [CLS] embedding from vision transformer}). The prototype replay in PASS and SSRE augments the class center (averaged over all sample prototypes of this class) to synthesize sample-level prototypes of old tasks and use them to maintain the old decision boundary. However, as mentioned in ~\cite{maprogressive}, the sample prototypes are not necessarily normally distributed. Due to the fact that DNNs tend to overfit to the training samples of the new task. These biased synthesized prototypes used for classifier replay may cause the classifier to remember wrong decision boundaries instead of preserving the initial old ones, leading to more severe forgetting. 

To solve the two problems described above, we propose a novel NECIL framework using the natural advantage of vision transformers: its patch representations. Our method consists of patch-level knowledge selection and prototype restoration. On the one hand, the vision transformer computes patch-level representations for each input image. According to the similarity between each patch and the [CLS] token embedding, the model perceives the relevance of each patch to the task and applies weighted knowledge distillation. For the foreground patches, the model tends to reduce the intensity of regularization and is encouraged to maintain plasticity on them. While the background patches may have little relevance to the task. Therefore the model can use them for preserving consistent representations over models of different historical versions. 
On the other hand, considering the hypothesis in PASS that the prototype has similar data distribution (normal distribution), we expand it into two steps. First, we compute the offset distance explicitly of each sample to its class center (\ie{ prototype}), and regularize them to maintain a task-agnostic data distribution. Then we use this property to restore prototypes of old tasks by current task prototypes and the old class prototypes. These operations mitigate the inaccurate restored old prototypes caused by adopting the Gaussian model in PASS and obtain more realistic prototypes for classifier replay, thus relieving the classifier bias and forgetting.

Our main contributions can be summarized as follows:

\textbf{(i)} We propose a novel vision transformer framework for NECIL tasks, in which patch-level knowledge selection can be naturally applied to achieve better trade-offs of network plasticity and stability. 

\textbf{(ii)} We adopt a novel prototype restoration strategy to generate more realistic synthesized prototypes for alleviating forgetting in the classifier.   

\textbf{(iii)} Extensive experiments performed on CIFAR-100, TinyImageNet, and ImageNet-Subset demonstrate the effect and superiority of our framework. Each component of our method can be easily applied to other related works with remarkable performance gain.

\section{Related Work}
\minisection{Incremental Learning}
Incremental learning involves learning sequential tasks with different knowledge, which has drawn much attention these years, including a variety of methods ~\cite{belouadah2021comprehensive,delange2021continual,liu2018rotate,liu2022long,zhou2023deep}. To overcome catastrophic forgetting due to insufficient access to old task data, iCaRL~\cite{rebuffi2017icarl} uses knowledge distillation between class logits from the current and old model. PODNet~\cite{douillard2020podnet} further applies distillation on each intermediate feature map in the backbone.

Recently, vision transformer has become popular in image classification and its derivation tasks like class incremental learning(CIL). DyTox~\cite{douillard2022dytox} replaces the backbone from CNN to the vision transformer and introduces task-relevant embeddings for adapting the model into incremental tasks. L2P and DualPrompt~\cite{wang2022learning,wang2022dualprompt} guide pre-trained transformer models through dynamic prompts, enabling them to learn tasks sequentially across different tasks. Besides, ~\cite{zhai2023masked} introduces the masked autoencoder to expand the replay buffer for class incremental learning.
In this paper, we rethink the characteristic of the vision transformer and expand it to NECIL as a new insight for alleviating catastrophic forgetting: fine-grained patch-level knowledge selection. 

\begin{figure*}[!t]
    \centering
    \includegraphics[scale=0.27]{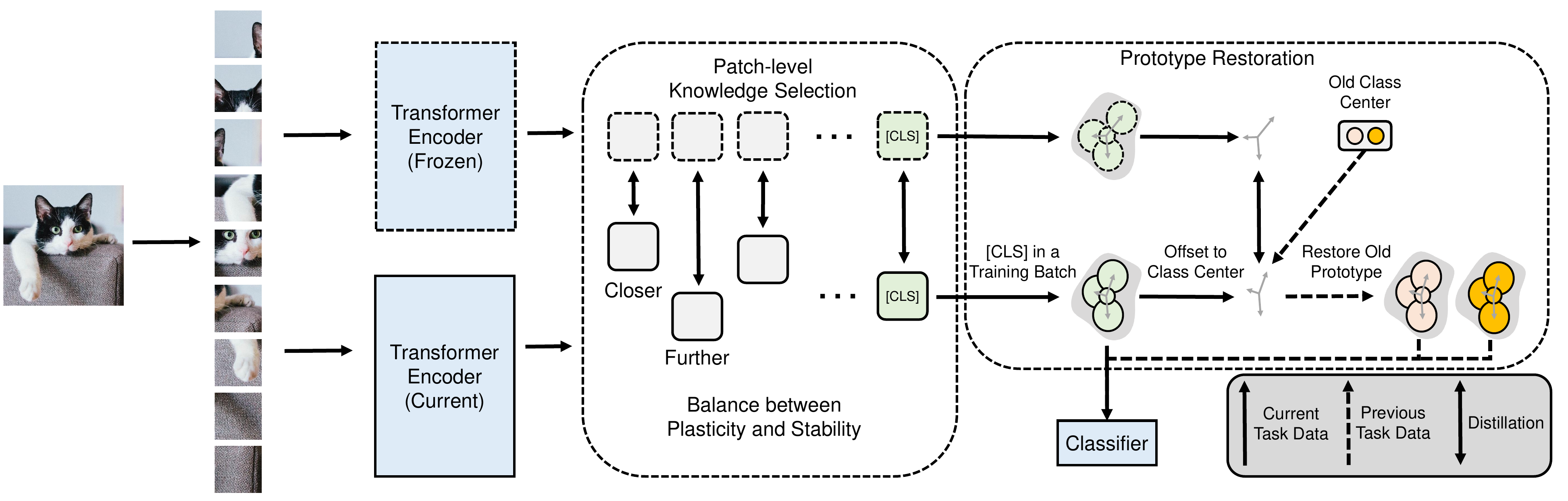}
	\caption{Illustration of the fine-grained knowledge selection and restoration framework of our method. The patch embeddings are regularized with different weights. We first train the current network to preserve similar prototype distribution with the old one, and restore old prototypes with old class center and current task prototypes. Two kinds of prototypes are sent to the classifier to reduce task bias. 
	}\label{fig:main}
  \vspace{-5mm}
\end{figure*}

\minisection{Non-exemplar Class Incremental Learning}

Non-exemplar class incremental learning (NECIL) is preferred in scenarios where training data is sensitive and cannot be stored long-term. DAFL employs a GAN to generate synthetic samples for past tasks, avoiding the need to store actual data~\cite{chen2019data}. DeepInversion, another NECIL method, generates images by inverting trained networks with random noise~\cite{yin2020dreaming}. SDC addresses semantic drift in training new tasks on old samples by estimating and utilizing prototypes drift~\cite{yu2020semantic}. Methods like PASS and IL2A~\cite{zhu2021prototype, zhu2021class} offer efficient NECIL by generating prototypes of old classes without retaining original images. SSRE introduces a re-parameterization method balancing old and new knowledge, and self-training leverages external data as an alternative for NECIL~\cite{zhu2022self, 21SC_WebSeg, yu2022self}.

The split prototypical replay in NECIL into task-relevant prototype center and task-agnostic prototype offset in our methods improves replay quality by first supervising the model to produce task-agnostic offsets, then using these to restore old class prototypes. This enhances prototype generation compared to the standard method in PASS~\cite{zhu2021prototype}.

\section{Method}

\subsection{Preliminaries}
\minisection{Problem definition and analysis}
Class-incremental learning sequentially learns different tasks. Each of these tasks does not have overlapping classes with previous ones. Let $t \in \{1,2,..T\}$ denotes the incremental learning tasks, where $T$ is the number of all tasks. The training data $D_t$ contains classes $C_t$ with $N_t$ training samples $\{(x_{t}^{i},y_{t}^{i})\}_{i=1}^{N_t}$. $x^i_t$ denotes images and $y_t^i \in C_t$ are their class labels.

Most deep networks of class-incremental learning can be split into two components: a feature extractor $F_\theta$ and a classifier $G_\phi$ which grows with each new task $t+1$ to include classes $C_{t+1}$.
The input $x$ is converted to a deep feature vector $z=F_\theta(x)\in \mathbb{R}^d$ by the feature extractor $F_\theta$, and then the unified classifier $G_\phi(z) \in \mathbb{R}^{|C_t|}$ is used to learn a probability distribution over classes $C_t$ for predicting $x$'s label.

Class-incremental learning requires the model to classify all learned samples from previous tasks at \textit{any} training task. In other words, The model should retain its ability to classify samples from classes belonging to tasks $t' < t$ while performing task $t$. Taking these requirements into account, non-exemplar class-incremental learning (NECIL) imposes an additional constraint that models must learn each new task without using any samples from previous tasks. Most related methods are supervised with a fundamental objective that minimizes a loss function $\mathcal{L}_{t}^{\text{CIL}}$ defined on current training data $D_t$:
\begin{equation}
  \begin{aligned}
    \label{eqn:cil-loss}
    \mathcal{L}_{t}^{\text{CIL}}(x, y) &=  \mathcal{L}_{t}(G_{\phi_t}(F_{\theta_t}(x)),y) .
  \end{aligned}
\end{equation}

\minisection{Vision transformer architecture}
DyTox~\cite{douillard2022dytox} has shown that vision transformer is effective for CIL with the dynamic task-relevant token which can be readily adapted across different tasks. In this paper, we discover one essential characteristic of vision transformer that can benefit CIL and adaptively alleviate forgetting in new tasks: its patch-level representation of images. The process of Visual Transformer is reviewed as follows:

Vision transformer first crops the input image $x$ into $K \times K$ non-overlapping patches, and we denote the number of patches in the full image $x$ as $N$. After this operation, the patches are mapped to a visual embedding of dimension $d$ with an MLP layer. The result of this, after concatenating with a class token [CLS] of shape $\mathbb{R}^{d}$, is a tensor of size $\mathbb{R}^{(N+1) \times d}$. 
After positionally encoding the original patch locations, the input is passed to the vision transformer Encoder. 
Each encoder transformer block has two sequential components: a self-attention layer and a feed-forward layer. LayerNorm is applied before each of these. 
These operation maintains the same shape of embedding, \ie{ $\mathbb{R}^{(N+1) \times d}$} and yields patch-level representations for each region.
The class token embedding from the processed result can be used for classification with a cross-entropy loss $\mathcal{L}_{t}^{\mathrm{CIL}}$. We adopt the linear classifier with softmax operation to predict the probability of each learned class. 

\subsection{Patch-level knowledge selection (PKS)}
When learning task $t$, only $D_t$ is available for the model. Vanilla knowledge distillation in previous NECIL methods like PASS and SSRE directly maintain the inter-task stability with current data. This operation does not consider the semantic-gap between current task samples and old ones, leading to suboptimal effects of alleviating forgetting on old tasks. To overcome this problem, we rethink the usage of knowledge distillation with patchified images. A natural idea is to assign different weights for each patch when applying knowledge distillation due to their respective importance to the classification task: patches in foreground regions often contain more task-relevant context compared to other patches in the background, which mostly have task-agnostic pixels with random information. 

Considering this bilateral strategy for balancing plasticity and stability of the model on $D_t$, we implement it in two steps: (a) define a metric for evaluating the relevance of each patch to the current task $t$, (b) apply patch-level knowledge selection on each patch with these patch-specific weights between the current model and old one. 

For simplicity without losing generality, we adopt the L2 distance between the embedding of [CLS] token $P_{t,cls}$ and each image patch $P_{t,i}$ to compute their importance to task $t$:
\begin{equation}
  \begin{aligned}
    \label{eqn:patchw}
    W_{i} &= \frac{1}{||P_{t,cls}-P_{t,i}||_2+\epsilon} .
  \end{aligned}
\end{equation}
We assign larger $W_i$ for patches that are closer to [CLS] token, and divide each $W_i$ by their maximum value for normalization and get $w_i$. The $\epsilon$ is set to $1e-8$ for avoiding zero value in the denominator. 
The patch-level knowledge selection $\mathcal{L}_{t}^{pks}$ is defined as:
\begin{equation}
  \begin{aligned}
    \label{eqn:pks}
    \mathcal{L}_{t}^{pks} &= \sum_{i=1}^{N} w_i ||P_{t,i}-P_{t-1,i}||_2 + ||P_{t,cls}-P_{t-1,cls}||_2,
  \end{aligned}
\end{equation}
 $P_{t,i}$ denotes the computed embedding for patch $i$ by the model of task $t$, and we compute the L2 distance of embeddings between the current model $F_{\theta_t}$ and old model $F_{\theta_{t-1}}$. 

\subsection{Prototype restoration (PR)}
We first describe the definition of prototype offset and class center. The feature extractor $F_{\theta}$ computes the representation $z \in \mathbb{R}^d$ from the input image $X_t$ of task $t$, which is used to predict the class label with the classifier $G_\phi$. As for the vision transformer, we adopt the [CLS] token as the image representation for classification. Let $N_{t,k}, X_{t,k}, \mu_{t,k}$ denote the number of samples, the image set, and the class center of class $k$ in task $t$. And $\mu_{t,k}=\frac{1}{N_{t,k}}\sum_{i=1}^{N_{t,k}} F_{\theta}(X_{t,k}^i)$, \ie{ averaged over all samples of this class}.

To introduce more realistic and forgetting-free sample-level prototypes for alleviating classifier bias, we restore the prototype of old tasks with their class-center and current samples. We divide our prototype replay into two steps: 1) introduce supervision to make these prototype offsets task-agnostic, and 2) use this characteristic to restore the old sample-level prototypes.

\begin{table*}[t]
	\centering
	\small
	\renewcommand{\arraystretch}{1.1}
  \setlength\tabcolsep{1.2mm}
	\resizebox{1\textwidth}{!}{
\begin{tabular}{c|cc|c|c|c|c|c|c|c|c|c|c|c|c|c|c}\hline
\multicolumn{3}{c|}{\textbf{Dataset}}&\multicolumn{6}{c|}{\textbf{CIFAR100} }&\multicolumn{6}{c|}{\textbf{TinyImageNet}} &\multicolumn{2}{c}{\textbf{ImageNet-Subset} }\\ \hline
	   \multicolumn{3}{c|}{\textbf{Setting}} & \multicolumn{2}{c|}{\textbf{5 tasks}} &  \multicolumn{2}{c|}{\textbf{10 tasks}} & \multicolumn{2}{c|}{\textbf{20 tasks}} & \multicolumn{2}{c|}{\textbf{5 tasks}} &  \multicolumn{2}{c|}{\textbf{10 tasks}} & \multicolumn{2}{c|}{\textbf{20 tasks}}&  \multicolumn{2}{c}{\textbf{10 tasks}} \\
\hline
\multicolumn{2}{c}{\textbf{Method}}&\textbf{Params(M)}&Avg$\uparrow$&Last$\uparrow$&Avg$\uparrow$&Last$\uparrow$&Avg$\uparrow$&Last$\uparrow$&Avg$\uparrow$&Last$\uparrow$&Avg$\uparrow$&Last$\uparrow$&Avg$\uparrow$&Last$\uparrow$&Avg$\uparrow$&Last$\uparrow$\\
\hline
\multirow{5}{*}{\textbf{\rotatebox{90}{~E=20}}}
&iCaRL-CNN\dag&11.2&51.07&40.12&48.66&39.65&44.43&35.47&34.64&22.31&31.15&21.10&27.90&20.46&50.53&41.08\\
&iCaRL-NCM\dag&11.2&58.56&49.74&54.19&45.13&50.51&40.68&45.86&33.45&43.29&33.75&38.04&28.89&60.79&51.90\\
&LUCIR\dag&11.2&63.78&55.06&62.39&50.14&59.07&48.78&49.15&37.09&48.52&36.80&42.83&32.55&66.16&56.21\\
&EEIL\dag&11.2&60.37&52.35&56.05&47.67&52.34&41.59&47.12&34.24&45.01&34.26&40.50&30.14&63.34&54.19\\
&RRR\dag&11.2&66.43&57.22&65.78&55.74&62.43&51.35&51.20&42.23&49.54&40.12&47.46&35.54&67.05&58.22\\
\hline
\multirow{6}{*}{\textbf{\textbf{\rotatebox{90}{~E=0}}}}
&LwF\_MC&14.5&45.93&36.17&27.43&50.47&20.07&15.88&29.12&17.12&23.10&12.33&17.43&\phantom{0}8.75&31.18&20.01\\
&EWC&14.5&16.04&\phantom{0}9.32&14.70&\phantom{0}8.47&14.12&\phantom{0}8.23&18.80&12.71&15.77&10.12&12.39&\phantom{0}8.42&-&-\\
&MUC&14.5&49.42&38.45&30.19&19.57&21.27&15.65&32.58&17.98&26.61&14.54&21.95&12.70&35.07&22.65\\
&IL2A&14.5&63.22&55.13&57.65&45.32&54.90&45.24&48.17&36.14&42.10&35.23&36.79&28.74&-&-\\
&PASS&14.5&63.47&55.67&61.84&49.03&58.09&48.48&49.55&41.58&47.29&39.28&42.07&32.78&61.80&50.44\\
&SSRE&19.4&65.88&56.33&65.04&55.01&61.70&50.47&50.39&41.67&48.93&39.89&48.17&39.76&67.69&57.51\\

&Ours&9.3&\textbf{68.17}&\textbf{59.02}&\textbf{70.13}&\textbf{57.90}&\textbf{66.86}&\textbf{54.25}&\textbf{54.88}&\textbf{44.97}&\textbf{52.72}&\textbf{43.35}&\textbf{51.68}&\textbf{41.94}&\textbf{70.18}&\textbf{61.42}\\
\hline
	\end{tabular} 
 }
 \vspace{-6pt}
	\caption{Average and last accuracy on  CIFAR100, TinyImageNet, and ImageNet-Subset under different numbers of tasks. Replay-based methods storing 20 exemplars from each previous class are denoted by $\dag$. The best overall results are in \textbf{bold}.}
	\label{tab:cmp}
   \vspace{-4mm}
  \end{table*}

  \begin{algorithm}[!t]
	\renewcommand{\algorithmicrequire}{\textbf{Input:}}
	\renewcommand{\algorithmicensure}{\textbf{Output:}}
	\caption{Pseudocode of Training Process} 
	\label{alg::vitne} 
	\begin{algorithmic}[1] 
		\Require 
		The number of tasks $T$, training samples $D_t = \{(x_i, y_i)\}_{i=1}^{N}$ of task $t$, class prototype $\mu_{t,k}$ of class $k$ in task $t$  (maintained during training), initial parameters $\Theta_0 = \{\theta_0, \phi_0\}$ containing parameters of a vision transformer feature extractor $F_{\theta_t}$, and a classifier $G_{\phi_t}$. CE denotes the cross entropy loss.
		\Ensure 
		Model $\Theta_T$
		
		\For {$t\in$ $\{1,2,...,T\}$}
		\State $\Theta_t$ ← $\Theta_{t-1}$
		\While {not converged}
        \State Sample $(x, y) \mbox{ from } D_t$
		\State $P_{t,i}, P_{t-1,i}$ ← $F_{\theta_t}(x), F_{\theta_{t-1}}(x)$
		\State $\mathcal{L}_{t}^{pks}$ ← Compute$(P_{t,i}, P_{t-1,i})$ in Eq.~(\ref{eqn:pks})
            \State $O_t, O_{t-1}$ ← $P_{t,y}-\mu_{t,y}, P_{t-1,y}-\mu_{t,y}$
		\State $\mathcal{L}_{t,pr}$ ← $\mathcal{L}_{mse}$($O_t, O_{t-1}$) in Eq.~(\ref{eqn:sploss})
            \State $F_{t_{old},y_{old}}$ ← $O_t$ + $\mu_{t_{old},y_{old}}$ in Eq.~(\ref{eqn:restore})
		\State $\mathcal{L}_{t}^{CIL}$ ←  $\mathcal{L}_{t}^{CE}$ $(G_{\phi_t}(F_{t,y},F_{t_{old},y_{old}}), y, y_{old})$   
		\State update $\Theta_t$ by minimizing $\mathcal{L}^{\mathrm{all}}_t$ from Eq.~(\ref{allloss})
		\EndWhile
		
		\EndFor

	\end{algorithmic} 
\end{algorithm}

\minisection{Supervision for task-agnostic prototype offset.} For the first part, we consider the model of the current and last task, \ie{ $F_{\theta_t}$ and $F_{\theta_{t-1}}$}, applying the offset regularization. 
Let $bs$ denote the batch size, we consider the training samples in a batch $(x_i^t,y_i^t), i=1,2,...,bs$, and randomly split them into two subsets $S_1$ and $S_2$ of the same size $\lfloor \frac{bs}{2} \rfloor$. Then, we compute the prototype offset of 
samples in the two subsets: $\{O_{t, i}=F_{\theta_t}(x_i^t)-\mu_{t,y_i^t}|i \in S_1\}$ and $\{O_{t-1, i}=F_{\theta_{t-1}}(x_i^t)-\mu_{t,y_i^t}|i \in S_2\}$. We adopt the old model $F_{\theta_{t-1}}$ for computing the prototype offset of subset $S_2$, which contains previous knowledge of the offset distribution. 
We randomly sample $\lfloor \frac{bs}{2} \rfloor$ pair of prototype offset from them, and minimize the mean square error between them:
\begin{equation}
\begin{aligned}
\mathcal{L}_{t}^{pr} &= \frac{1}{sz}\sum_{(i_k,j_k)\in Idx} \mathcal{L}_{mse}(O_{t, i_k},O_{t-1, j_k}),
\label{eqn:sploss}
\end{aligned}
\end{equation}
%
where $sz =\lfloor \frac{bs}{2} \rfloor$, $Idx = \{(i_1,j_1),...,(i_{sz},j_{sz})\} (i_k\in S_1, j_k\in S_2)$, $O_{t,i_k}$ denotes the $i_k$-th prototype offset from $S_1$, and $O_{t-1,j_k}$ has the similar meaning.

\minisection{Restoration of old task prototype.} We use the prototype offset of current sample $x_i^t$ to restore prototype from old tasks:
\begin{equation}
\begin{aligned}
 F_{t_{old},k_{old}} &=  \mu_{t_{old},k_{old}} + (F_{\theta_{t}}(x_i^t)-\mu_{t,y_i^t}).
\label{eqn:restore}
\end{aligned}
\end{equation}
The second term in Eq.~\ref{eqn:restore} is the computed offset from the current sample. The sample $(x_i^t,y_i^t)$ is randomly selected within the current batch. It can be fused in Eq.~\ref{eqn:cil-loss} as follows,

\begin{equation}
  \begin{aligned}
    \label{eqn:cil-loss}
    \mathcal{L}_{t}^{\text{CIL}} &=  \mathcal{L}_{t}^{CE}(G_{\phi_t}(F_{t,y},F_{t_{old},y_{old}}), y, y_{old}) ,
  \end{aligned}
\end{equation} 
where $\mathcal{L}_{t}^{CE}$ is the cross entropy (CE) loss. An overall algorithm is illustrated in Alg. ~\ref{alg::vitne}.

\subsection{Learning Objective}
The overall learning objective combines the classification loss, sample prototype consistency loss, and patch-level knowledge selection:
\begin{equation}
  \begin{aligned}
  \label{allloss}
    \mathcal{L}^{\text{all}}_t &= \mathcal{L}^{\mathrm{CIL}}_t + \lambda_{pks} \mathcal{L}^{\text{pks}}_t + \lambda_{pr} \mathcal{L}^{\text{pr}}_t. 
  \end{aligned}
\end{equation}

\section{Experiments}
\minisection{Datasets.}
We conduct experiments on three datasets: CIFAR100, TinyImageNet, and ImageNet-Subset, as commonly used in previous works. For each experiment, we first select partial classes from the dataset as the base task and evenly split the remaining classes into each sequential task. This process can be represented with $F+C \times T$, where $F$, $C$, $T$ denote the number of classes in the base task, the number of classes in each task, and the number of tasks. For CIFAR100 and ImageNet-Subset, we adopt three configurations: 50 + 5 $\times$ 10, 50 + 10 $\times$ 5, 40 + 20 $\times$ 3. For TinyImageNet, the settings are: 100 + 5 $\times$ 20, 100 + 10 $\times$ 10, and 100 + 20 $\times$ 5. 

\minisection{Comparison Methods.}
We compare our method with other non-exemplar class incremental learning methods: SSRE~\cite{zhu2022self}, PASS~\cite{zhu2021prototype}, IL2A~\cite{zhu2021class},  EWC~\cite{kirkpatrick2017overcoming}, LwF-MC~\cite{rebuffi2017icarl}, and MUC~\cite{liu2020more}. We also compare with several exemplar-based methods like iCaRL (both nearest-mean and CNN)~\cite{rebuffi2017icarl}, EEIL~\cite{castro2018end}, and LUCIR~\cite{hou2019learning}. 

\minisection{Implementation Details.}
As for the structure of the vision transformer, we use 5 transformer blocks for the encoder and 1 for the decoder, which is much more lightweight than the original version of Vit-Base. All transformer blocks have an embedding dimension of 384 and 12 self-attention heads. We train each task for 400 epochs. After task $t$, we save one averaged prototype (class center) for each class. We set $\lambda_{pks}$ and $\lambda_{pr}$ to 10 in experiments. We report three common metrics of CIL task: the average and last top-1 accuracy after learning the last task on all learned tasks and average forgetting for all classes learned up to task $t$. we use $Acc_i$ to denote the accuracy over all learned classes after task $i$. Then the average accuracy is defined as $Avg\_acc=\frac{\sum_{i=1}^T Acc_i}{T}$. And the last accuracy is $Acc_T$. Let $a_{m,n}$ denotes the accuracy of task $n$ after learning task $m$. The forgetting measure $f_k^i$ of task $i$ after learning task $k$ is computed as $f_k^i=\max_{t\in1,2,...,k-1}(a_{t,i}-a_{k,i})$. The average forgetting $F_k$ is defined as $F_k=\frac{1}{k-1}\sum_{i=1}^{k-1}f_k^i$. 
We perform three runs of all experiments and report the mean performance.

\begin{figure*}[t]
    \centering
    \includegraphics[scale=0.26]{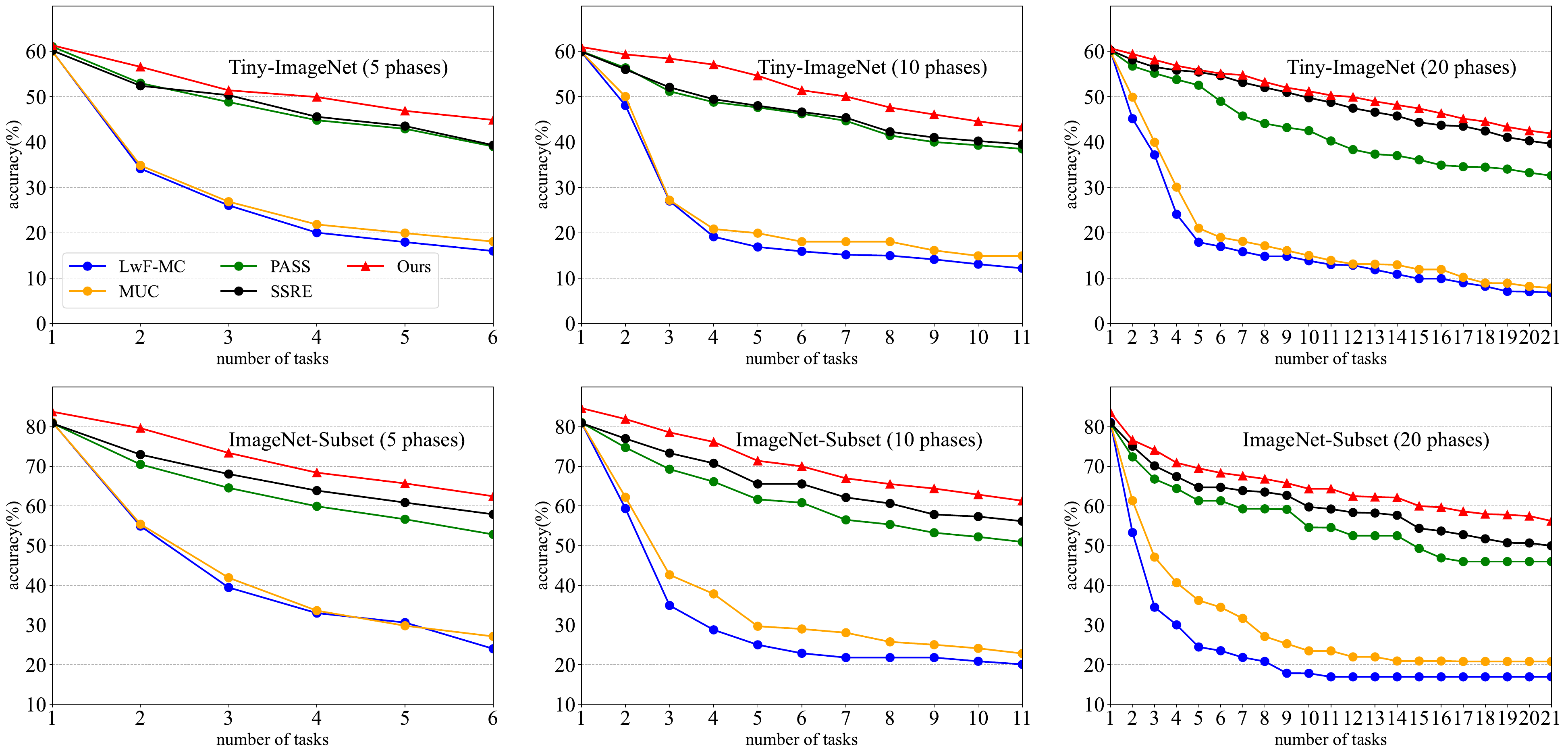}
	\caption{ 
	Results on TinyImageNet and ImageNet-Subset for different numbers of tasks. Our method outperforms others, especially on longer task sequences (i.e. more, but smaller, tasks). 
	}\label{fig:curve}
  \vspace{-5mm}
\end{figure*}

\subsection{Comparison with the State-of-the-art}

  \begin{table}[t]
	\centering
	\small
	\renewcommand{\arraystretch}{1.1}
  \setlength\tabcolsep{1.2mm}
	
 \resizebox{0.48\textwidth}{!}{
\begin{tabular}{c|ccc|ccc|ccc}\hline
\multicolumn{1}{c|}{\textbf{Dataset}}&\multicolumn{3}{c|}{\textbf{CIFAR100}} &\multicolumn{3}{c|}{\textbf{TinyImageNet}} &\multicolumn{1}{c}{\textbf{ImageNet-Subset} } \\ \hline
	   \textbf{Method} & \textbf{5 tasks} &  \textbf{10 tasks} & \textbf{20 tasks} & \textbf{5 tasks} &  \textbf{10 tasks} & \textbf{20 tasks} &   \textbf{10 tasks} \\
\hline
LwF\_MC&44.23&17.04&55.46&54.26&54.37&63.54&56.07\\
EWC&60.17&62.53&63.89&67.55&70.23&75.54&71.97\\
MUC&40.28&47.56&52.65&51.46&50.21&58.00&53.85\\
IL2A&23.78&30.41&30.84&25.43&28.32&35.46&32.43\\
PASS&25.20&30.25&30.61&18.04&23.11&30.55&26.73\\
SSRE&18.37&19.48&18.37&\textbf{9.17} &14.06& 14.20&23.22\\
\hline
Ours&\textbf{16.36}&\textbf{17.16}&\textbf{16.61}&11.45&\textbf{12.21}&\textbf{12.82}&\textbf{18.39}\\
\hline
	\end{tabular} 
	
	}
 \vspace{-6pt}
	\caption{Comparisons of the average forgetting with other methods. Experiments are conducted on CIFAR100, TinyImageNet, and ImageNet-Subset with 5, 10 and 20 task. 
 }
	\label{tab:avgforget}
  \end{table}
\begin{table}[tp]
	\centering
	\small
	\setlength\tabcolsep{1.3mm}
	\renewcommand{\arraystretch}{1.3}
	
	\begin{tabular}{ccc|ccc}
	\hline
	   Method &PKS&PR &5 tasks&10 tasks&20 tasks \\ \hline
    PASS&-&-&55.67&49.03&48.48\\
	   	   Baseline (ViT) &&& 56.44 & 51.90 & 51.20 \\
& \checkmark && 58.27 & 56.82 & 52.87 \\
&& \checkmark & 57.78 & 54.61 & 51.51 \\
& \checkmark & \checkmark & 59.02 & 57.90 & 54.25 \\
	   \hline

	\end{tabular}
 \vspace{-6pt}
	\caption{Ablative experiments on each component of our proposed method. Experiments are conducted on CIFAR-100 and we report the top-1 accuracy in \%. 
	We use PKS, PR to denote patch-level knowledge selection and prototype restoration, respectively. 
	}
	\label{tab:abl}
  \vspace{-5mm}
  \end{table}

In Table~\ref{tab:cmp}, we compare our methods with several non-exemplar and exemplar-based methods.
For the non-exemplar setting, we discover that our method outperforms all previous related methods in different data split setting (5/10/20 tasks) on all three datasets. 
Take the result of 20 tasks as an example, we surpass the best non-exemplar methods SSRE by 3.31\% on 20 tasks setting of CIFAR100 (Last Avg accuracy). In addition, our method even achieves higher accuracy than all exemplar-based methods using stored samples for alleviating forgetting. 
This phenomenon remains the same in datasets with a larger resolution, \ie{ TinyImageNet and ImageNet-Subset}. For the average forgetting reported in Table~\ref{tab:avgforget}, our method outperforms most non-exemplar-based methods. The gap (up to 4.17 \%) is clearer on ImageNet-Subset dataset. It demonstrates the superior performance of our method from another perspective during the incremental training process.
We also show the dynamic accuracy curves in Figure~\ref{fig:curve}, the results show that the proposed method (in red) has less rate of decline during all the training phases.


\subsection{Ablation Study}

\newcommand{\addts}[1]{\includegraphics[width=0.195\linewidth]{src/fig/patchattn/#1}}
\begin{figure*}[t]
  \centering
  \small
  \setlength\tabcolsep{0.2mm}
    \renewcommand\arraystretch{0.6}
    \begin{tabular}{cccccc}
    \rotatebox{90}{\scriptsize{~~~~~~~~~~~~~~~~~~~~Image}}&\addts{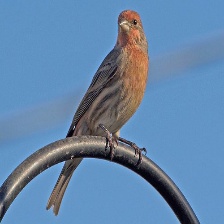}&\addts{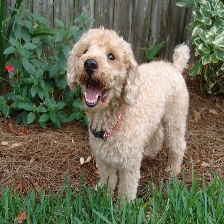}&\addts{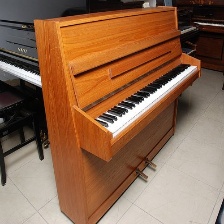}&\addts{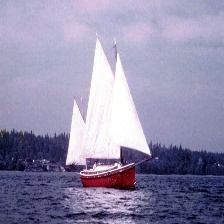}&\addts{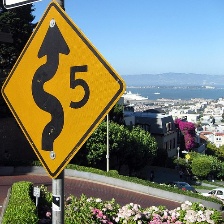}\\
    \rotatebox{90}{\scriptsize{~~~~~~~~~~~~~~~Patch KD}}&\addts{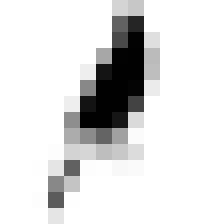}&\addts{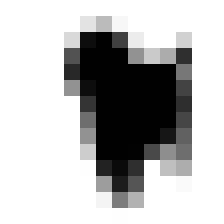}&\addts{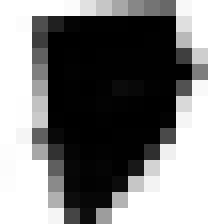}&\addts{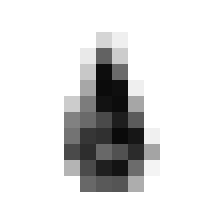}&\addts{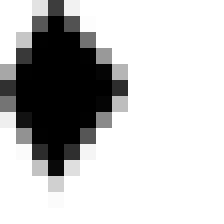}\\
    &Task 1&Task 3&Task 5&Task 7&Task 9\\
    \end{tabular}
    \vspace{-6pt}
		\caption{
	Visualization of our patch-level knowledge selection and its comparison with vanilla knowledge distillation. Patches with white color denote larger weights of distillation for preserving stability. 
		}
		\label{fig:patch}
   \vspace{-5mm}
\end{figure*}
\minisection{Each component.}
Our method consists of two components: patch-level knowledge selection and prototype restoration. We analyze the effect of each aspect in Table~\ref{tab:abl}. 
One baseline is trained with vanilla knowledge distillation and prototype augmentation in PASS. We consider vision transformer (ViT) training as another baseline.
We could discover that: (a) The patch-level knowledge selection significantly improves the performance by 3.71\%. (b) The prototype restoration also yields some gain by providing the more realistic prototype replay. (c) These two factors can collaborate with each other and achieve higher performance. It validates the significance of both our patch-level knowledge selection and prototype restoration for non-examplar class incremnetal learning setting.

We introduce two modules in PASS to conduct experiments with this backbone: prototype augmentation and self-supervision, as listed in the second row of Table~\ref{tab:abl}. For example, our framework has similar or slightly higher performance with the original network in PASS (\ie{ResNet18}) comparing results in the first two rows, showing that vision transformer could serve as a new baseline for further study. This also demonstrates the effect of our proposed two modules on vision transformer, instead of a stronger baseline. Since the Dynamic Structure Reorganization (DSR) in SSRE is designed on the convolutional layer, we do not conduct experiments on it. 
\begin{table}[t]
	\centering
	\small
	\renewcommand{\arraystretch}{1.1}
  \setlength\tabcolsep{1.2mm}
	
	\resizebox{.48\textwidth}{!}{
	\begin{tabular}{c|ccc|ccc|cccccccc} 
\hline
	   \multirow{2}{*}{Setting} &\multicolumn{3}{c|}{$N$=10}&\multicolumn{3}{c|}{$N$=20}&\multicolumn{3}{c}{$N$=50} \\ 
   &Avg$\uparrow$&Last$\uparrow$&$F\downarrow$&Avg$\uparrow$&Last$\uparrow$&$F\downarrow$&Avg$\uparrow$&Last$\uparrow$&$F\downarrow$\\ \hline
   DyTox&75.47&62.10&15.43&75.10&59.41&21.60&73.89&57.21&24.22\\
   Ours&78.35&66.47&13.12&77.63&63.90&15.76&76.52&61.45&20.23\\
\hline
	\end{tabular} 
 }
 \vspace{-6pt}
	\caption{Results applied to Dytox on CIFAR-100 in average accuracy (\%), last accuracy (\%), and forgetting $F$ (\%) on 10-, 20- and 50-task scenarios.}
	\label{tab:dy}
  \end{table}

\begin{table}[tp]
	\centering
	\small
	\setlength\tabcolsep{1.3mm}
	\renewcommand{\arraystretch}{1.3}
	\begin{tabular}{c|cccc}
	\hline
	   $W_i$ &5 tasks&10 tasks&20 tasks  \\ \hline
    1 & 56.18 & 51.99 & 50.53 \\
$||P_{t,cls}-P_{t,i}||_2$ & 53.81 & 49.78 & 48.31 \\
Eq.~\ref{eqn:patchw} (Ours) & 59.02 & 57.90 & 54.25 \\
	   \hline
	\end{tabular}
 \vspace{-6pt}
	\caption{Experiments are conducted on CIFAR-100 and we report the top-1 accuracy in \%. The first row set all $W_i$ to 1, and the second row uses $W_i$ proportional to the patch embedding's distance to [CLS] embedding. 
	}
	\label{tab:pksw}
  \vspace{-5mm}
  \end{table}
  
\minisection{More study of patch-level knowledge selection.}
Since our patch-level knowledge selection is a simple but effective extension of vanilla knowledge distillation based on vision transformer, we apply it on DyTox and exemplar-based tasks for verifying its general application with more vision transformer methods and problem settings. 
For each experiment in Table~\ref{tab:dy}, we store 20 exemplars of each learned class following DyTox for fair comparisons. The vanilla knowledge distillation is replaced with our patch-level knowledge selection.
We observe that the proposed method outperforms the original DyTox with a large gain in terms of Average/ Last average accuracy and Forgetting. It further proves the effectiveness of our insight on fine-grained patch distillation. Compared with vanilla knowledge distillation, our method preserves the knowledge by adaptive weights with different patches, offering more flexibility for the model to balance between plasticity and stability.

Furthermore, given that our patch-level knowledge selection approach utilizes varying weights for each patch embedding during the distillation process, comparative experiments were also conducted to evaluate the effectiveness of this strategy in two distinct settings.
The first one sets all distillation weight $W_i$ to 1, and the second one computes it by $W_{i} = ||P_{t,cls}-P_{t,i}||_2$ to replace Eq.~\ref{eqn:patchw}. 
According to results of the first setting in Table~\ref{tab:pksw}, we found that placing the same weights to the distillation of all patches results in worse performance than Ours (the third row). We assume that this strict restriction preserves more information about the previous task while hurting the learning ability of the current task. Then assigning more weights for distillation on embeddings further to the [CLS] embedding, which is opposite to our method, leads to inferior results than the baseline. The results of both settings demonstrate the importance of preserving plasticity on task-relevant patches and stability on task-agnostic patches.

\begin{table}[tp]
	\centering
	\small
	\setlength\tabcolsep{1.3mm}
	\renewcommand{\arraystretch}{1.3}
	
	\begin{tabular}{c|ccc}
	\hline
	   Method&5 tasks&10 tasks&20 tasks  \\ \hline
	   	   PASS&55.67&49.03&48.48\\
PASS+PR &57.45&52.30&51.22\\
	   \hline
	   SSRE&56.33&55.01&50.47\\
SSRE+PR&58.02&57.10&53.28\\
	   \hline

	\end{tabular}
 \vspace{-6pt}
	\caption{Experiments are conducted on CIFAR-100 and we report the top-1 accuracy in \%, demonstrating the effect of integrating prototype restoration on PASS and SSRE, which is denoted by `PR'.
	}
	\label{tab:spa}
  \vspace{-5mm}
  \end{table}

\minisection{Prototype restoration on other methods.}
The prototype restoration is designed to produce more realistic synthesized prototype of old tasks, compared with prototype augmentation in PASS and SSRE adopting a gaussian model for generating prototype offset of each class. We apply this method into PASS and SSRE with CNNs  as backbone in Table~\ref{tab:spa}. Our prototype restoration method yields a substantial increase in accuracy, ranging from 2 to 3 points across all task settings. This highlights the importance of using data from similar distributions for replay, not only for transformer-based backbones but also for CNNs.

\minisection{Visualization of patch-level knowledge selection.}
We visualize some examples and demonstrate the actual weights applied by our patch-level knowledge selection across different tasks in Figure~\ref{fig:patch}. The images are picked from ImageNet-Subset with the 10-task setting.
 Instead of using the same weight on every patch like vanilla knowledge distillation, our method can adaptively select some background patches for preserving stability, while offering more plasticity on foreground patches to learn task-relevant knowledge.

\section{Conclusions}

This paper introduces a novel framework using vision transformers for non-exemplar class incremental learning (NECIL) to reduce catastrophic forgetting and classifier bias. It utilizes patch embeddings for a balanced stability-plasticity trade-off in different regions and employs unique strategies for task-relevant and task-agnostic areas. A new prototype restoration module is also introduced to preserve the decision boundary of old tasks without inducing classifier bias. The framework demonstrates superior performance over existing methods on various NECIL benchmark datasets, providing a potential baseline for future research.

\minisection{Acknowledgements} This work is funded by  
NSFC (NO. 62225604, 62206135, 62202331), 
and the Fundamental Research Funds for the Central Universities 
(Nankai Universitiy, 070-63233085). 
Computation is supported by the Supercomputing Center of Nankai University.

\bibliography{aaai24}

\end{document}